# tidyHEBO: Improving the Robustness of Heteroskedastic Evolutionary Bayesian Optimization

Zhukov L.A.[1,2], Shaburova E.V.[2], Antonets D.V.[1,2]

[1] *AI Center MSU, Lomonosov Moscow State University, Moscow, Russia*

[2] *MSU Institute for Artificial Intelligence, Lomonosov Moscow State University, Moscow, Russia*

## Abstract

Bayesian optimization is widely used for expensive black-box optimization problems. While methodological advances often focus on individual components, practical performance depends on how these components work together. We present tidyHEBO, a Bayesian optimization model. tidyHEBO jointly learns output warping and Gaussian-process parameters, evaluates acquisition functions on the original objective scale, and preserves promising novel candidates throughout evolutionary search. On the Olympus benchmark, tidyHEBO ranked first among the evaluated methods, achieving the best average ranks for typical optimization quality, worst-tail outcomes, and run-to-run variability. It also obtained results comparable to specialized optimizers on Needle-in-a-Haystack problems. On the Bayesmark benchmark, tidyHEBO also demonstrated strong results in hyperparameter tuning. Across these benchmarks, tidyHEBO combines strong optimization results with robustness across diverse black-box problems.

## Introduction

Bayesian optimization (BO) is a sample-efficient strategy for optimizing expensive black-box objectives. A probabilistic surrogate model approximates the objective, and an acquisition function (AF) selects the next point by balancing exploitation of promising regions against exploration of uncertain ones. BO has been widely used in materials discovery [1,2], drug discovery [3], reaction and catalyst optimization [4,5], and particle-accelerator tuning [6]. In these settings, BO has been shown to accelerate experimental optimization, produce solutions comparable in quality to those proposed by domain experts [4], and identify high-performing solutions that would be difficult to discover by intuition alone [7]. Its use in demanding scientific optimization problems has therefore expanded rapidly. In practice, however, performance depends on how well the surrogate assumptions match the geometry and noise structure of the objective.

A common surrogate in BO is a Gaussian process (GP) with homoskedastic observation noise. This assumption is violated when measurement variance changes systematically across the design space or with the magnitude of the response. Such mean-dependent or relative-error structures occur in many measurement settings [8,9]. For example, quantitative NMR measurements used to estimate reaction yields and selectivity can exhibit concentration-dependent uncertainty [10]. If heteroskedasticity is ignored, GP uncertainty estimates can be poorly calibrated, reducing the efficiency of acquisition-based search [11,12] (Figure 1A, B).

Another common assumption in GP-based BO is kernel stationarity. A stationary kernel assumes that covariance depends on relative displacement rather than absolute position in the design space. This can be inadequate when the objective varies smoothly in some regions but rapidly in others [13,14] (Figure 1C, D). Similar behavior can arise in systems whose response changes across time or operating regimes [15,16]. A stationary GP may then provide inaccurate estimates of the response surface or its uncertainty [13,17], which in turn can reduce optimization efficiency.

HEBO [12] is an influential optimizer developed to address related difficulties in hyperparameter optimization (HPO). The original study emphasized heteroskedasticity and non-stationarity and combined input/output warping with evolutionary search over multiple acquisition criteria under small evaluation budgets. HPO and general scientific optimization share several structural features, including limited budgets, mixed parameter scales, non-convex response surfaces, and limited prior knowledge. HEBO has accordingly been used outside HPO, including as a baseline in antibody design [18] and Needle-in-a-Haystack (NiaH) problems [19]. However, HEBO was developed and primarily validated in HPO settings [12,20], leaving open how well its implementation choices transfer to broader scientific optimization problems.

In this work, we present tidyHEBO, a single-objective Bayesian optimizer built on BoTorch and designed for robust, general-purpose optimization. The method retains the useful high-level ideas of warped surrogate modeling and multi-acquisition Pareto search, but revises how these transformations, acquisition evaluation, and candidate-selection stages are implemented and coupled.

We make three main contributions. First, tidyHEBO jointly fits a regularized Yeo-Johnson output transformation with the GP, evaluates acquisition criteria on the original objective scale, and enforces novelty within a cumulative Pareto search. Reproducing HEBO in BoTorch [21] provided the starting point for identifying implementation choices that motivated these changes. Second, we evaluate tidyHEBO (without benchmark-specific hyperparameter tuning) across single-objective Olympus tasks [22,23] and quantify both performance and robustness using median nAUC, worst-tail CVaR_nAUC, and run-to-run IQR_nAUC. Third, to broaden the evaluation, we test tidyHEBO on synthetic functions, NiaH problems [19], adaptive batch optimization, and all 108 Bayesmark HPO tasks [20]. Across these problems, using the same default configuration, tidyHEBO performs strongly and stays close to HEBO on Bayesmark, supporting its use as a general-purpose optimizer.

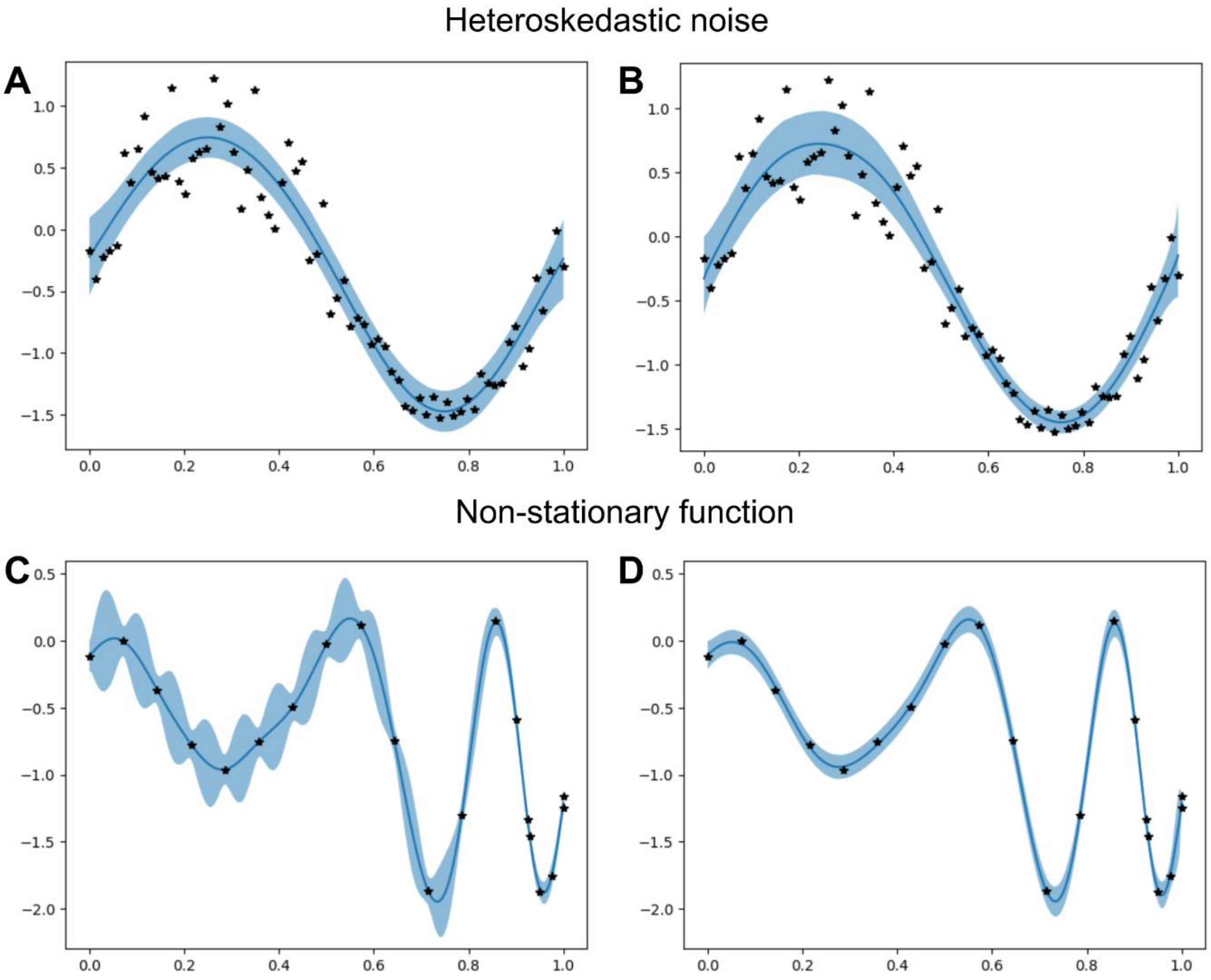

**Figure 1 | Illustrations of heteroskedasticity and non-stationarity.** A heteroskedastic objective modeled using (A) a conventional GP and (B) an output-warped GP. A non-stationary objective modeled using (C) a GP with a stationary kernel and (D) a GP with a stationary kernel applied to warped inputs.

# Results

## Methodological design of tidyHEBO

tidyHEBO retains three high-level elements of HEBO – input and output warping, multiple acquisition criteria, and evolutionary Pareto search – but redesigns how these components are implemented and coupled. Reconstructing the method in BoTorch exposed implementation choices that materially affect surrogate fitting, acquisition evaluation, and candidate selection. Table 1 summarizes the resulting differences, and Figure 2 shows the tidyHEBO workflow.

| Category | Model part | HEBO | tidyHEBO |
|---|---|---|---|
| **Framework** | Base BO framework | Standalone implementation | BoTorch |
| **GP model** | Kernel type | Matérn 3/2 + Linear | Matérn 5/2 |
| | Prior distributions | Task-specific | General-purpose |
| **Warping** | Input warping | Kumaraswamy-CDF | Kumaraswamy-CDF |
| | Gated output warping | Box–Cox or Yeo–Johnson; unwarped fallback | Yeo–Johnson or standardization fallback |
| | Output warping training | Separately from GP | Jointly optimized with GP |
| **Acquisition function** | Acquisition criteria | [log(EI), PI, UCB] | [EI, PI, UCB] |
| | Acquisition perturbation | Explicit via hyperparameter | No explicit perturbation |
| | Acquisition evaluation space | Warped objective space | Original objective space |
| | Sequential acquisition evaluation | Direct evaluation in warped space | Analytic PI; quadrature for EI and UCB when applicable; QMC fallback |
| **Optimization of acquisition functions** | Genetic algorithm | NSGA-II | NSGA-III |
| | Novelty filtering | *After* Pareto front search | Enforced *during* Pareto front search |
| | Pareto front selection | From the *last* generation only | Cumulative across *all* generations |

**Table 1 | Comparison of the main methodological components of HEBO and tidyHEBO.**

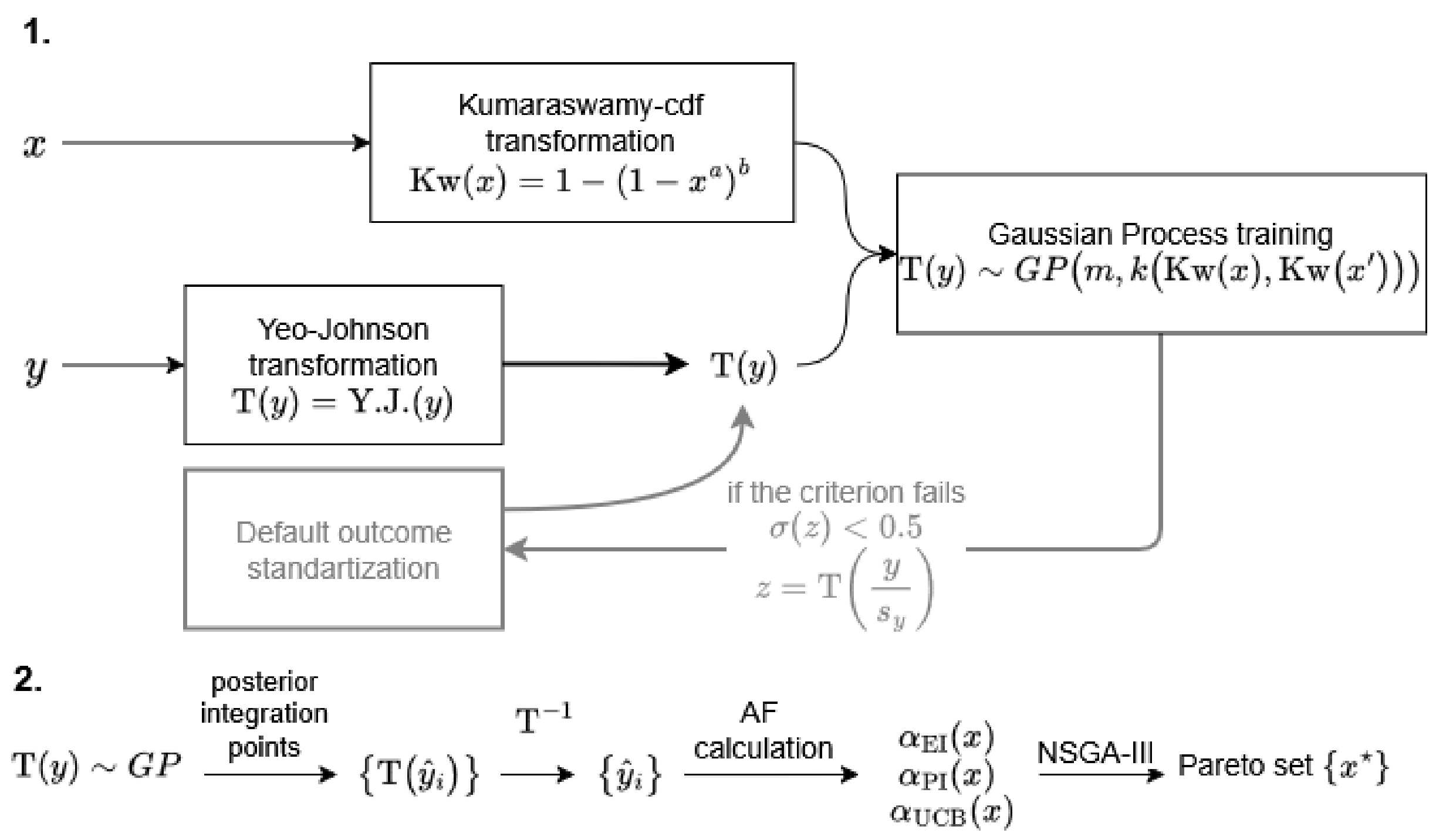

**Figure 2 | Workflow of tidyHEBO. 1.** Input and output warping during Gaussian-process surrogate training and inference; **2.** Acquisition function evaluation and candidate selection.

Output warping is one such difference. In HEBO, Box-Cox or Yeo-Johnson transformation parameters are estimated separately from the GP and therefore without the covariance structure subsequently assumed by the surrogate. tidyHEBO instead optimizes the Yeo-Johnson parameter jointly with the GP hyperparameters, making output warping part of the same probabilistic model (Figure 2, step 1). The illustrative comparison in Supplementary Figure 1 shows how this joint training changes the fitted mean and uncertainty.

A second difference concerns the objective scale used for acquisition evaluation. HEBO evaluates acquisition criteria directly on warped model outputs. For a nonlinear transformation, EI and UCB are not generally invariant, so rankings on the warped scale can differ from those on the original objective scale. In tidyHEBO, posterior integration points are mapped back to the original scale before acquisition evaluation (Figure 2, step 2). PI can be evaluated directly from the latent Gaussian distribution because the transformation is strictly increasing, whereas EI and the moments used for UCB require integration. For $0 \leq \lambda \leq 2$, these one-dimensional integrals are evaluated using Gaussian quadrature; quasi-Monte Carlo (QMC) integration is used as a fallback when quadrature is not applicable (Supplementary Note 4).

Two further changes concern the Pareto search over acquisition criteria. First, tidyHEBO imposes novelty as a feasibility constraint during evolutionary optimization, whereas HEBO checks candidate novelty only after constructing the Pareto front; previously evaluated points can therefore participate in non-dominated sorting and suppress useful novel candidates before the final filter is applied. Second, tidyHEBO retains candidates evaluated throughout the search and determines the final non-dominated set from their cumulative history, whereas HEBO constructs the returned Pareto set only from the single final generation, where any good solutions that emerged during evolution may already have been lost.

Finally, tidyHEBO uses EI on its original scale when combining it with PI and UCB in the multi-objective search. When output warping is active and one-dimensional quadrature is applicable, EI in the original objective space is evaluated directly using deterministic Gauss–Legendre quadrature. In cases where the standard BoTorch EI formulation is used, the numerically stable logarithmic expected improvement formulation [24] is evaluated and exponentiated before entering the acquisition triplet. This preserves a consistent EI scale across the acquisition paths while benefiting from stable intermediate calculations in the logarithmic formulation. Exponentiation can still underflow for extremely small EI values. Moreover, a strictly increasing transformation from EI to log(EI) preserves exact Pareto dominance; its practical effects can arise through numerical behavior and scale-dependent evolutionary search operations.

Not all differences between HEBO and tidyHEBO summarized in Table 1 should be interpreted in the same way. The use of BoTorch, a Matérn-5/2 kernel, and NSGA-III reflects deliberate design choices rather than modifications motivated by implementation-specific behavior. For further details, see Methods.

## tidyHEBO performs strongly on synthetic test functions

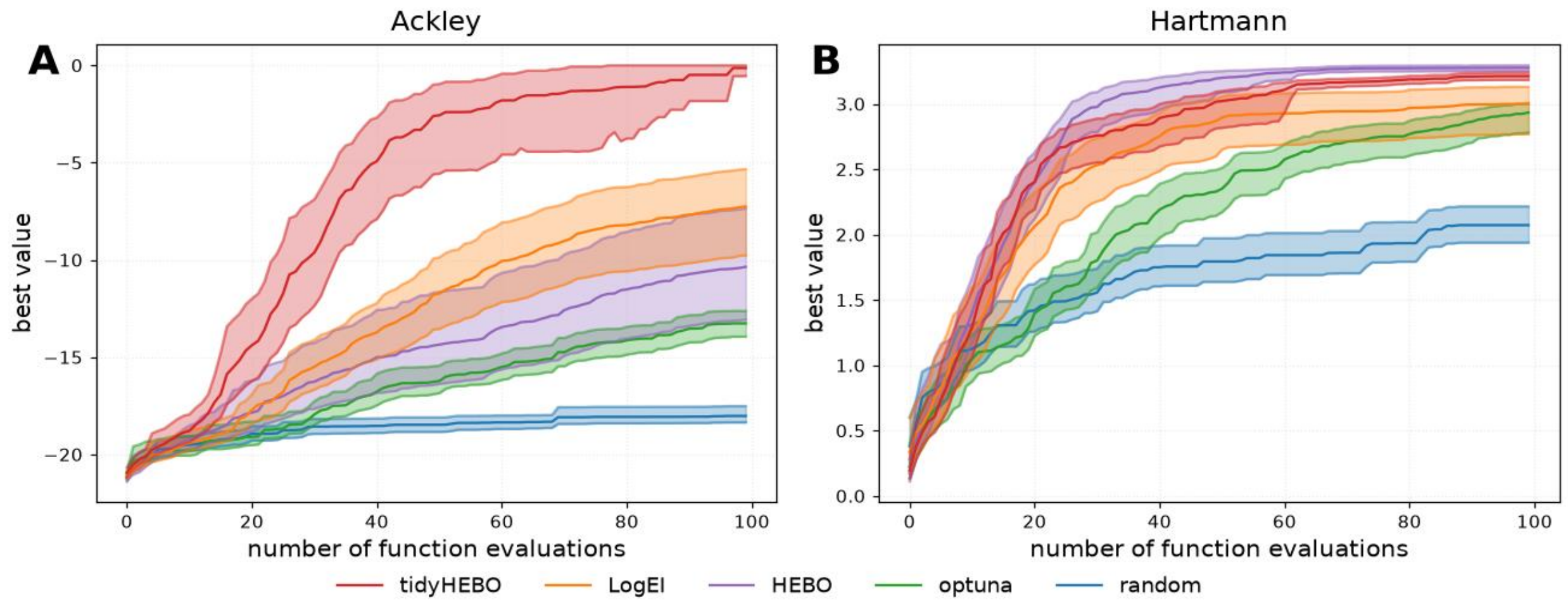

**Figure 3 | Optimization results on synthetic functions: Ackley (A), Hartmann (B).** Curves show the mean best maximization-oriented objective value obtained up to each optimization step versus the number of function evaluations with 95% bootstrap confidence intervals. Results are aggregated across 30 optimization runs with 100 function evaluations each.

We first evaluated tidyHEBO on the six-dimensional Ackley and Hartmann functions [25] to probe behavior on multimodal landscapes with multiple local optima and different response scales. We compared tidyHEBO with random search, the default TPE sampler in Optuna [26], a conventional GP using logarithmic expected improvement as its acquisition function (denoted LogEI), and HEBO. To characterize both optimization quality and robustness, we calculated the median normalized area under the regret curve (nAUC), CVaR_nAUC for poor-tail outcomes, and IQR_nAUC for run-to-run variability (see Methods).

tidyHEBO showed the lowest median nAUC on Ackley (Figure 3A) and remained competitive with HEBO on Hartmann (Figure 3B; for evaluation details, see Supplementary Note 5). On Ackley, tidyHEBO achieved the lowest median nAUC (25.5) and CVaR_nAUC (74.0), indicating better performance and poor-tail robustness than the other evaluated methods (Supplementary Table 1). On Hartmann, HEBO achieved the best median nAUC (14.2 versus 18.0), whereas tidyHEBO showed greater consistency, with the best IQR_nAUC (5.7) and second-best CVaR_nAUC (33.8; Supplementary Table 2).

## tidyHEBO ranks first for performance and robustness on the Olympus benchmark

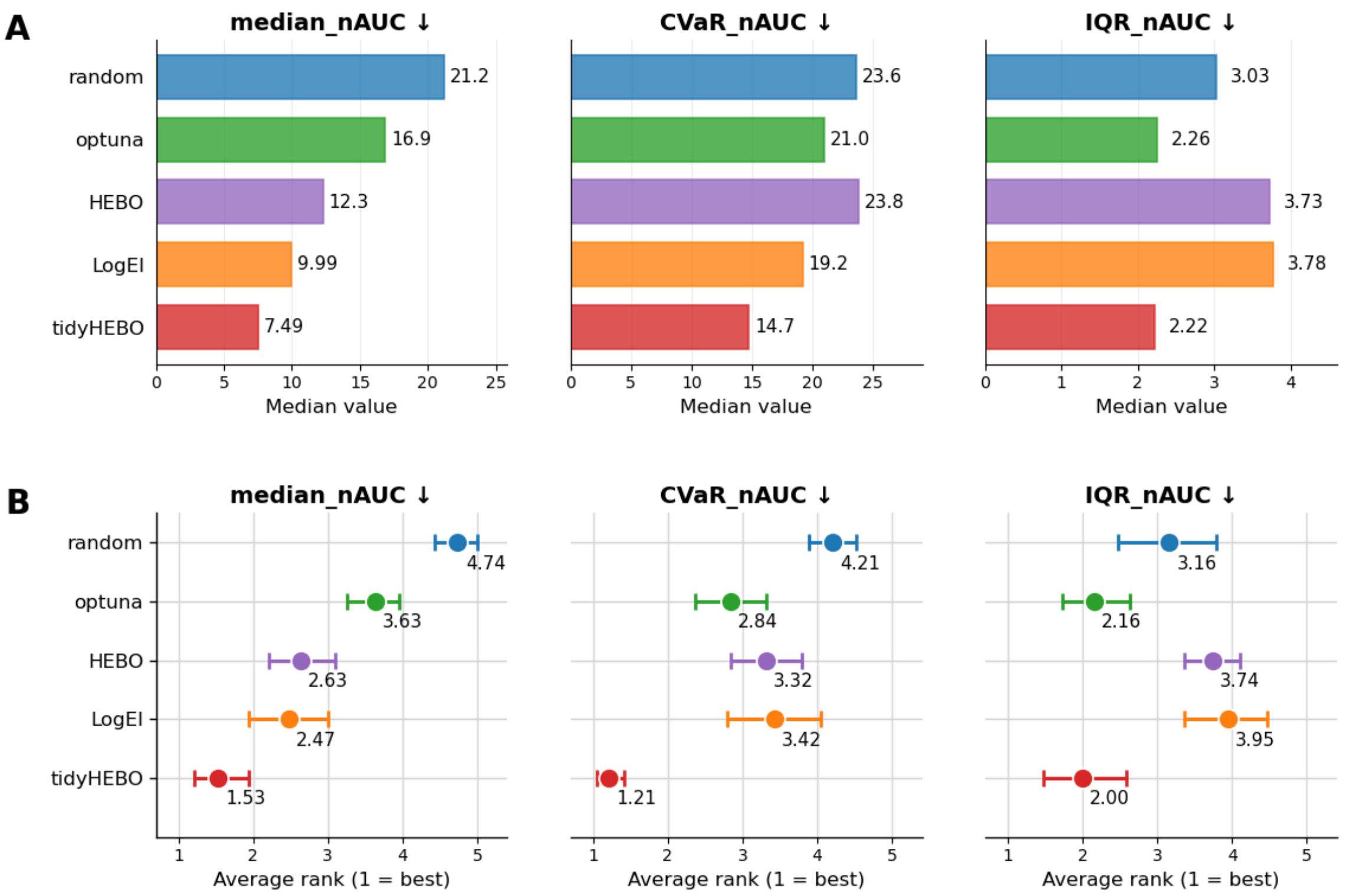

**Figure 4 | Performance and robustness on Olympus. (A)** Medians across tasks of median nAUC, CVaR_nAUC, and IQR_nAUC for tidyHEBO and the baselines; lower values are better. **(B)** Average task-level ranks for the same metrics, with rank 1 indicating the best method. Horizontal error bars show 95% bootstrap confidence intervals obtained by resampling tasks. The parametrization of simplex-constrained problems is described in Supplementary Note 3.

We evaluated tidyHEBO on single-objective Olympus tasks using three complementary metrics that capture both optimization quality and robustness. Median nAUC summarizes typical anytime performance, IQR_nAUC measures variability across independent runs, and CVaR_nAUC summarizes the worst-performing tail of the run distribution. Considering the three metrics jointly is important because low variability alone does not imply good optimization: a method can be consistently mediocre, whereas a robust optimizer should combine strong typical performance with limited dispersion and a limited frequency or severity of poor outcomes.

All three metrics are regret-based and lower values therefore indicate better performance. tidyHEBO showed the best aggregate performance across all three metrics (Figure 4). When task-level values were summarized using the median across Olympus tasks, tidyHEBO achieved median nAUC = 7.49, CVaR_nAUC = 14.7, and IQR_nAUC = 2.22, the lowest values among the compared methods (Figure 4A). Despite normalization of objective values to a common min–max utility scale, metric values still varied substantially across tasks (Supplementary Figure 2), reflecting differences in optimization difficulty and response-surface structure. We therefore used the median rather than the mean for cross-task aggregation to reduce sensitivity to unusually difficult problems.

Average rank was used as the primary cross-task comparison (Figure 4B). tidyHEBO ranked first for all three metrics, with average ranks of 1.53 for median nAUC, 1.21 for CVaR_nAUC, and 2.00 for IQR_nAUC. Its advantage was largest for typical and worst-tail performance. For IQR_nAUC, tidyHEBO was close to Optuna (2.16), but this similarity did not extend to CVaR_nAUC, for which Optuna ranked 2.84. The comparatively favorable IQR_nAUC of random search further illustrates why dispersion should not be interpreted independently of optimization quality: consistently poor outcomes can also yield a narrow interquartile range. Notably, tidyHEBO was evaluated using the same default configuration across all Olympus tasks, without task-specific hyperparameter tuning.

## Adaptive batching exposes a controllable feedback-round versus performance trade-off in tidyHEBO

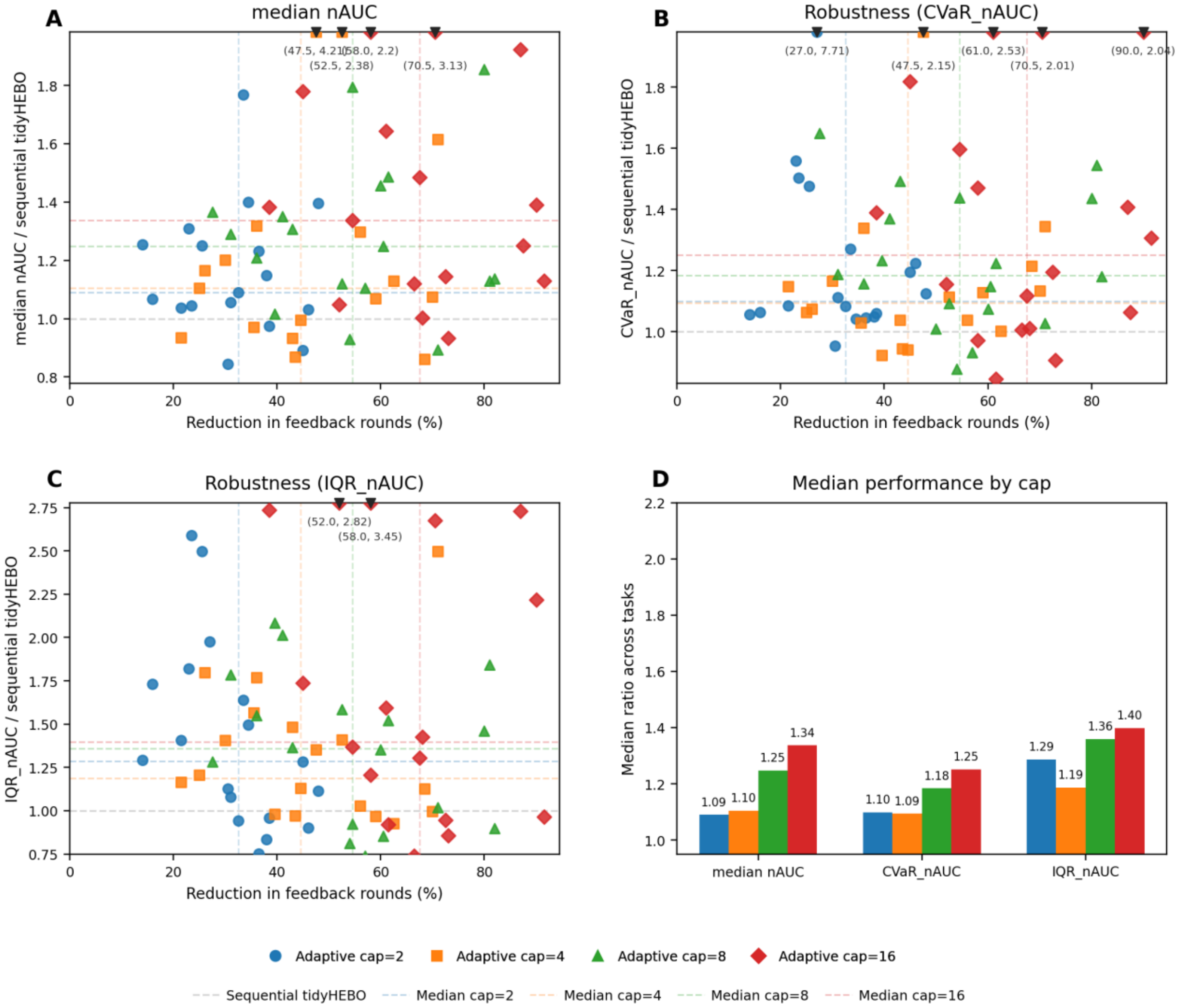

**Figure 5 | Adaptive-batching trade-offs on Olympus. (A–C)** Each point represents one task at a batch cap of 2, 4, 8, or 16. The x-axis shows the reduction in feedback rounds relative to sequential tidyHEBO. The y-axis shows the ratio to the sequential baseline for (A) median nAUC, (B) CVaR_nAUC, or (C) IQR_nAUC; a ratio of 1 indicates the baseline. Colored horizontal and vertical dashed lines show the median metric ratio and median reduction in feedback rounds, respectively. Black downward-pointing triangles mark points above the plotted range. (**D**) Median metric ratios across tasks for each cap; bar labels give these medians.

We next evaluated whether tidyHEBO could reduce the number of feedback rounds by returning multiple candidates at once, using Pareto sets limited to adaptive batch caps of 2, 4, 8, and 16. Larger caps generally produced greater reductions in feedback rounds at the cost of higher nAUC (Figure 5A). Across tasks, the median nAUC ratio relative to sequential tidyHEBO increased from 1.09 at cap = 2 to 1.10, 1.25, and 1.34 at caps of 4, 8, and 16, respectively (Figure 5D).

The median CVaR_nAUC ratios were 1.10, 1.09, 1.18, and 1.25 at caps of 2, 4, 8, and 16, respectively; the corresponding IQR_nAUC ratios were 1.29, 1.19, 1.36, and 1.40 (Figure 5D). Larger caps generally reduced the number of feedback rounds while worsening median nAUC. The robustness metrics did not worsen monotonically with the cap: cap 4 had lower median CVaR_nAUC and IQR_nAUC ratios than cap 2. In these experiments, caps of 2 or 4 provided a favorable balance between parallelization and sequential optimization performance.

## tidyHEBO remains competitive on Needle-in-a-Haystack problems

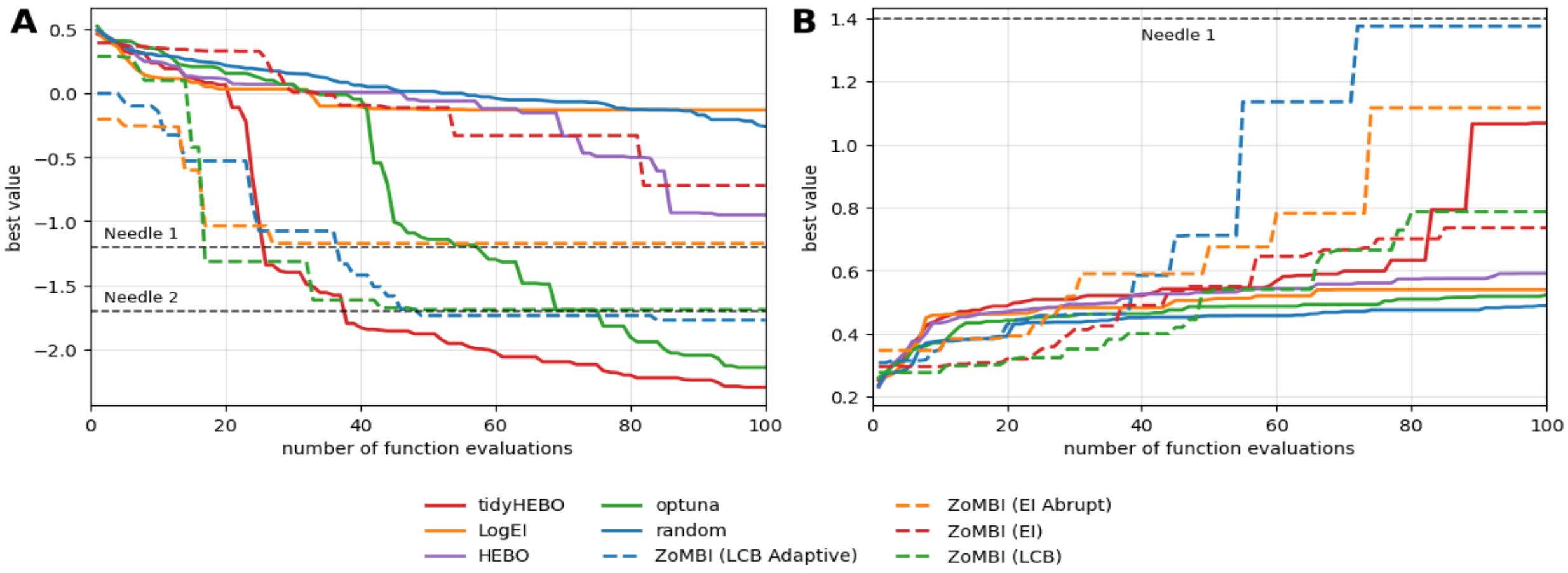

**Figure 6 | Results on Needle-in-a-Haystack problems.** Evaluation of tidyHEBO on NiaH problems from [1**9]:** (A) finding materials with highly negative Poisson's ratio (poisson_ratio emulato**r),** (B) identifying rare materials with the highest thermoelectric figure of merit (thermoelectrics emulator). Curves for the optimizers evaluated here show the median best-so-far objective value over 30 runs. Results for the ZoMBI methods were taken from the original publication [19]; see Supplementary Note 5 for evaluation details.

The results on Ackley and Olympus motivated additional stress tests on the NiaH emulators introduced in [19]. These problems contain extremely localized high-performing regions embedded in otherwise weakly informative search spaces. The Ackley function shares one NiaH-like feature – a narrow basin around the global optimum – but its symmetry and synthetic construction differ substantially from realistic materials-design landscapes.

On the negative-Poisson's-ratio task (Figure 6A), tidyHEBO substantially outperformed random search and was comparable with the dedicated ZoMBI methods [19], whereas LogEI performed no better than random search. HEBO improved on random search but approached only the weaker ZoMBI variants. On the more challenging thermoelectrics task (Figure 6B; maximization), which contains a narrower high-performing region, tidyHEBO remained competitive with the ZoMBI collection, whereas HEBO and LogEI showed substantially weaker performance.

NiaH problems can be viewed as a challenge for stationary surrogate models, where the objective changes sharply within a narrow region and remains largely uninformative elsewhere. They probe whether an optimizer can identify highly localized high-performing regions in a search space that provides limited information away from the optimum. The results therefore broaden the empirical evaluation of tidyHEBO beyond canonical synthetic functions and the Olympus benchmark.

## tidyHEBO retains competitive HPO performance on Bayesmark

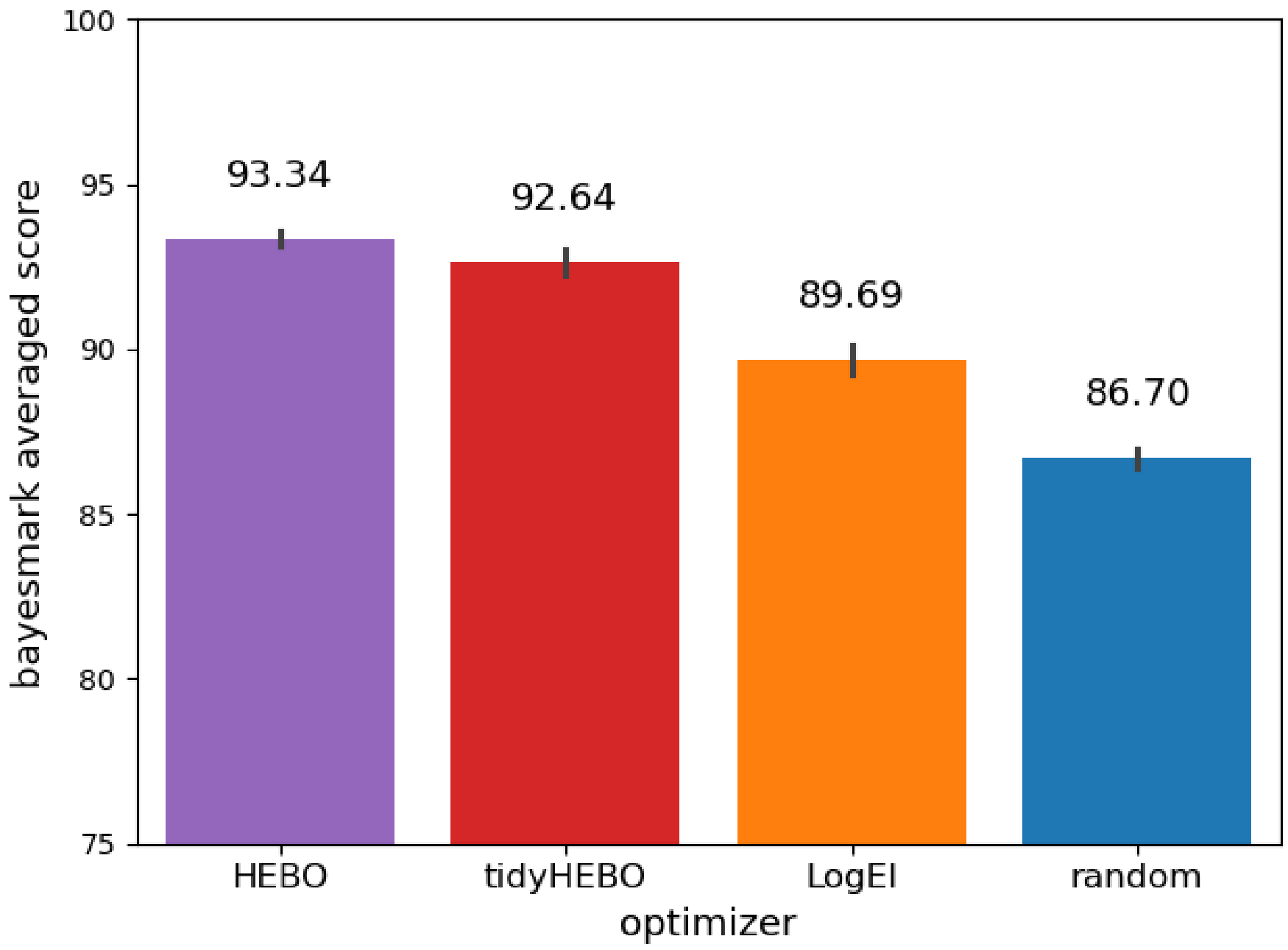

**Figure 7 | Results on the Bayesmark benchmark.** Evaluation of tidyHEBO on Bayesmark [20]. Each optimizer was tested on all 108 Bayesmark problems. Bars show the mean aggregated leaderboard score [20] across 20 repeated evaluations; error bars indicate 95% bootstrap confidence intervals across repeats. Experimental and scoring details are provided in Supplementary **Note 5.**

Finally, we evaluated tidyHEBO on Bayesmark [20], the HPO benchmark that motivated HEBO (Figure 7). In our sequential protocol with a single suggestion per step, HEBO achieved an average leaderboard score of 93.34, close to the published BBO Challenge score of 93.51 [20], which was obtained under a different batching and task selection protocol. This numerical agreement provides a useful consistency check on the evaluation pipeline rather than a strict replication. tidyHEBO achieved 92.64 with a confidence interval overlapping that of HEBO, compared with 89.69 for LogEI and 86.70 for random search. Thus, the modifications that extend tidyHEBO beyond HPO preserve performance close to HEBO on this benchmark under our protocol.

# Conclusions

We developed tidyHEBO as a general-purpose Bayesian optimizer implemented in BoTorch, combining a jointly trained GP and output transformation, input warping, acquisition evaluation on the original objective scale, and constrained cumulative Pareto search. tidyHEBO retains the multi-acquisition strategy of HEBO while substantially revising both the probabilistic model and candidate selection procedure. The strongest evidence for its practical performance comes from the Olympus benchmark: across 19 single-objective tasks, tidyHEBO achieved the best average rank among the evaluated methods for optimization quality, worst-tail performance, and run-to-run variability. These results were obtained using a single default optimizer configuration applied throughout the study without benchmark-specific hyperparameter tuning.

Results on additional problem classes support the generality of this behavior. The synthetic tests provided an initial performance assessment, although the large performance advantage of tidyHEBO on Ackley was a notable result. On the NiaH problems, tidyHEBO approached the performance of specialized ZoMBI methods while outperforming the conventional BO baselines. On Bayesmark, tidyHEBO achieved a leaderboard score of 92.64 compared with 93.34 for HEBO, while exceeding LogEI and random search, providing an additional check that the method remains competitive in the HPO setting for which HEBO was originally developed. Adaptive batching further showed that moderate caps of 2 or 4 can substantially reduce the number of sequential feedback rounds while preserving much of the performance of sequential tidyHEBO; the median nAUC increased by only 9% and 10%, respectively. Together, these results support tidyHEBO as a robust and reproducible general-purpose optimizer across diverse optimization problems.

# Methods

## Sequential single-objective optimization task

The single-objective optimization task is formulated as

$$\max_{x \in X} f(x), f: X \to \mathbb{R},$$

where $X$ is a bounded feasible set (*design space*), $x$ is a d-dimensional vector of decision variables, and $f(x)$ is the objective function to be optimized.

$$y = f(x) + \varepsilon.$$

Here, y denotes the observed response and ε denotes observation noise. A conventional unwarped GP assumes additive Gaussian noise with constant variance. In tidyHEBO, this Gaussian-noise assumption applies to the transformed response when output warping is active; the corresponding noise in the original response space is generally non-Gaussian and response-dependent.

tidyHEBO is formulated as a maximization algorithm. Minimization objectives are transformed to the corresponding maximization problem before optimization.

## Gaussian process surrogate model

A Gaussian process is specified by a mean function m and a covariance function k. For any finite input set $X = (x_1, \dots, x_n)$, the vector of latent function values has distribution $\mathbf{f} \sim \mathcal{N}(\mathbf{m}, K)$, where $m_i = m(x_i)$ and $K_{ij} = k(x_i, x_j)$ [11]. For an output-warped GP, this Gaussian assumption applies to the latent function in the transformed response space.

The covariance function (kernel) specifies how function values at different inputs are related. Kernel $k(x, x')$ is stationary if it depends only on the displacement between inputs; an isotropic stationary kernel depends only on their Euclidean distance $k(x, x') = k(||x - x'||)$.

The kernel used in this work is an automatic relevance determination (ARD) Matérn 5/2. For two input vectors x and x', the scaled distance is

$$r(x, x') = \sqrt{(x - x')^\top \Theta^{-2} (x - x')}$$

Here, Θ is the diagonal matrix of ARD length scales, and the covariance is

$$k(x, x') = (1 + \sqrt{5}r + \frac{5r^2}{3}) exp(-\sqrt{5}r)$$

Thus, each input dimension has its own length scale $l_j$. The kernel is used directly, without an additional learned output-scale parameter.

Conditioning the GP on the training observations gives the posterior mean and latent-function variance below. Let $\mathbf{k}(x) = \left(k(x_1, x), \dots, k(x_n, x)\right)^\top$ and let t denote the vector of training targets on the scale modeled by the GP, with mean vector m. Then

$$\mu(x) = m(x) + \mathbf{k}(x)^\top (K + \sigma_n^2 I)^{-1} (\mathbf{t} - \mathbf{m}).$$
$$\sigma^2(x) = k(x, x) - \boldsymbol{k}(x)^\top (K + \sigma^2{}_n I)^{-1} \boldsymbol{k}(x)$$

Here, $\sigma_n^2$ is the learned observation-noise variance on the modeled response scale. The variance above excludes observation noise; the predictive variance for a future noisy target additionally includes $\sigma_n^2$. When output warping is active, these Gaussian posterior quantities are on the transformed scale; original-scale moments must be obtained after inverse transformation.
For a comprehensive treatment of Gaussian-process regression, see Rasmussen and Williams [11].

## Acquisition functions

Acquisition functions are criteria to select the next point to evaluate. Their role is to balance exploration and exploitation.

Improvement is a measure of potential utility defined as:

$$I(x) = (\nu - \tau)\mathbb{I}(\nu > \tau),$$

Here, τ is the best observed objective value so far, and ν is the objective-scale predictive random variable at x. For an unwarped GP, ν has a Gaussian predictive distribution. For an output-warped GP, it is obtained by inverse transformation of a latent Gaussian variable and is generally non-Gaussian.

Probability of improvement (PI) measures the probability of exceeding the best observed objective value:

$$\alpha_{PI} = \mathbb{P}(\nu > \tau)$$

Expected improvement (EI) is the expected value of the improvement:

$$\alpha_{EI} = \mathbb{E}[I(x)]$$

Another acquisition function is upper confidence bound (UCB). It is defined as:

$$\alpha_{UCB} = \mu(x) + \sqrt{\beta}\sigma(x)$$

Here, μ(x) and σ(x) denote the predictive mean and standard deviation on the original objective scale, and β controls the exploration–exploitation trade-off.

Selecting the next point to evaluate is usually done by maximizing the acquisition function.

## Evaluation of optimization performance and robustness

Here, robustness means reproducibility of optimization performance under stochastic variation in initial designs, random seeds, surrogate fitting, and acquisition function optimization. For each optimization problem (benchmark task) $p$, optimizer $a$, repeated run $r$, and evaluation step $t$, let $y_{p,a,r,t}$ denote the raw objective value, and let $B$ denote the total experimental budget.

For each problem p, the best and worst reference values, $y_p^*$ and $y_p^w$, were estimated using differential evolution [27,28]. The utility function for each problem is defined as follows:

$$u_p(y) = \frac{y - y^w{}_p}{y^*{}_p - y^w{}_p}[maximization]\ ;\ u_p(y) = \frac{y^w{}_p - y}{y^w{}_p - y^*{}_p}[minimization]$$

Thus, $u_p(y) = 0$ corresponds to the estimated worst reference and $u_p(y) = 1$ to the estimated best reference. Values outside the interval [0, 1] are not clipped; a run can therefore have negative regret if it exceeds the estimated best reference. For each run, the incumbent utility and normalized simple regret are

$$U_{p,a,r}(t) = \max_{1 \le i \le t} u_p(y_{p,a,r,i});\quad R_{p,a,r}(t) = 1 - U_{p,a,r}(t).$$

## Median normalized regret AUC

Run-level normalized regret AUC (nAUC) is the trapezoidal area under the complete normalized best-so-far regret trajectory. Here, normalization applies to objective values through the utility function; the area is not divided by the evaluation horizon B − 1, so its scale depends on the budget.

$$nAUC_{p,a,r} = \sum_{t=1}^{B-1} \frac{R_{p,a,r}(t) + R_{p,a,r}(t+1)}{2}$$

The median is calculated over run-level nAUC values:

$$median\, nAUC_{p,a} = median_r(nAUC_{p,a,r})$$

## Interquartile range of normalized regret AUC

Run-to-run dispersion of complete optimization performance was quantified using the interquartile range of nAUC (IQR_nAUC).

$$IQR\ nAUC_{p,a} = Q_{p,a}(0.75) - Q_{p,a}(0.25)$$

Here, $Q_{p,a}(q)$ denotes the empirical q-quantile of run-level nAUC values for problem p and optimizer a.

## Worst-tail normalized regret AUC

To quantify poor optimization outcomes, we calculated the empirical upper-tail mean of nAUC at or above its 90th percentile, used here as an estimate of conditional value at risk (CVaR_nAUC). Let $q_{0.9} = Q_{p,a}(0.9)$; then

$$CVaR_{0.9}\ nAUC_{p,a} = mean\{nAUC_{p,a,r} : nAUC_{p,a,r} \geq q_{0.9}\}$$

Because larger nAUC values are worse, CVaR_nAUC summarizes the upper, poor-performance tail of the run distribution.

## tidyHEBO implementation and model specification

**Framework:** tidyHEBO is implemented in BoTorch [21]. This provides tested GP and acquisition-function components, makes numerical choices explicit, and facilitates reuse and extension within current BO workflows.

**Surrogate model choice:** we retain a GP surrogate because the low-evaluation-budget setting considered here requires both a flexible response-surface model and posterior uncertainty for acquisition-based decisions.

**Kernel:** we use an ARD Matérn-5/2 kernel as a general-purpose default [4,29,30]. The Matérn-5/2 covariance corresponds to twice mean-square-differentiable latent functions and provides an intermediate smoothness assumption between rougher Matérn kernels and the squared-exponential kernel [17,31].

General-purpose BoTorch priors are used instead of task-specific priors. For each ARD length scale, $\log(l_j) \sim \mathcal{N}(\sqrt{2} + 0.5\log(d), 3)$, where the second normal-distribution parameter is the variance and d is the input dimension [32]. Length scales have a lower bound of 0.025.

The GP uses a learned constant mean function $m$. Observation-noise variance on the modeled response scale is inferred jointly with the remaining GP parameters using a log-normal prior with log-space mean −4 and standard deviation 1, and a lower noise-variance bound of $10^{-4}$ [32].

To increase flexibility under spatially varying characteristic scales and response-scale heterogeneity, tidyHEBO applies adaptive input and output transformations.

**Input warping:** as in HEBO, tidyHEBO uses Kumaraswamy-CDF input warping. A stationary kernel in the warped coordinates can induce a non-stationary covariance in the original coordinates.

**Gated output warping:** in HEBO's reference implementation [33] output warping is implemented using a Box-Cox transformation [34,35] when all observed outcomes are positive and a Yeo-Johnson [36] transformation otherwise, with the untransformed response retained as a fallback if the transformation fails. Both transformations are applied using sklearn.preprocessing.power_transform, which standardizes the transformed values by default; therefore, the subsequent check of their standard deviation does not normally act as an effective variance-based gate after a successful **transformation.**

In tidyHEBO, the observed outcomes are first rescaled by their sample standard deviation $s_y$. The transformed response is $z = T_\lambda(y/s_y)$ and its inverse is $y = s_y T_\lambda^{-1}(z)$. For an effectively constant response, $s_y$ is set to one to avoid numerical instability. The Yeo–Johnson parameter is fitted jointly with the GP. After fitting, the sample standard deviation of the transformed training targets provides a gating criterion. If $\mathrm{sd}(z) < 0.5$, the nonlinear transformation is rejected and the GP is refitted using standard BoTorch outcome standardization. Thus, the alternatives are Yeo–Johnson warping and a standardization fallback.

Box–Cox requires strictly positive responses, whereas the Yeo–Johnson transformation is defined for all real-valued inputs. We regularize its learnable parameter using a normal prior, $\lambda \sim \mathcal{N}(1, 0.5^2)$, restricted to $[-3,3]$. For λ outside [0, 2], the transformation's range is bounded on one side, requiring special handling of latent Gaussian predictions outside the domain of the inverse transformation (see Supplementary Note 4).

Output-warping training: tidyHEBO fits the transformation jointly with the GP using the warped-GP approach [37]. By contrast, separate maximum-likelihood fitting of Box–Cox or Yeo–Johnson under an independent univariate Gaussian model ignores the input-dependent covariance structure used by the GP. Joint fitting instead evaluates the GP marginal likelihood on the transformed training vector and includes the Jacobian term and parameter priors. Supplementary Figure 1 illustrates the difference in fitted predictions. The transformation is therefore selected for the full GP model rather than solely to improve marginal univariate normality. The combined objective is derived in Supplementary Note 1.

**Acquisition triplet:** the three acquisition objectives are EI, PI, and UCB. When output warping is active and one-dimensional quadrature is applicable, EI in the original objective space is evaluated directly using deterministic Gauss–Legendre quadrature. Otherwise, when the standard BoTorch EI path is used, the numerically stable logarithmic expected improvement formulation [24] is evaluated and exponentiated before being combined with PI and UCB in the multi-objective acquisition search.

**Acquisition perturbation:** HEBO uses Gaussian perturbations of acquisition-function values to reduce sensitivity to surrogate-model misspecification. This introduces an additional noise-level hyperparameter. Unlike HEBO, tidyHEBO does not add explicit random perturbations to acquisition-function values. When output warping is used, the integrals defining acquisition functions in the original objective space are evaluated using deterministic Gaussian quadrature where applicable, while quasi-Monte Carlo estimation is retained as a fallback. In the default implementation, 512 quadrature nodes or 4096 quasi-Monte Carlo samples are used, respectively.

**Acquisition-evaluation space:** HEBO evaluates acquisitions using the Gaussian posterior in the transformed response space. tidyHEBO instead evaluates original-scale utility after inverse transformation, $Y(x) = s_y T_\lambda^{-1}[Z(x)]$. This follows the broader principle of evaluating utility under the induced distribution of the quantity of interest, as in composite-function BO [38]. Nonlinear warping can alter EI and moment-based UCB rankings. PI is invariant under a strictly increasing transformation when the improvement threshold is transformed consistently. See Supplementary Note 4 for numerical details and the inverse-domain qualification.

**Genetic algorithm:** the three acquisition criteria define a multi-objective candidate-search problem. tidyHEBO uses the pymoo implementation [39] of NSGA-III [40] instead of the NSGA-II procedure [41] used in HEBO. NSGA-III uses reference directions to maintain diversity across the non-dominated candidate set.

**Novelty filtering:** in the reference HEBO implementation, candidate novelty is checked after Pareto-front construction, and random points may be sampled if the final front contains no unobserved candidates. Because previously evaluated points participate in non-dominated sorting, they can dominate otherwise useful novel candidates before the final filter is applied. tidyHEBO instead imposes novelty as a feasibility constraint during evolutionary search, so previously evaluated points are excluded from the admissible Pareto set.

**Pareto-set selection:** HEBO constructs the returned Pareto set from the final evolutionary generation. tidyHEBO instead retains nondominated candidates across generations and selects from the resulting cumulative Pareto archive.

# Data and code availability

Code for tidyHEBO is available at https://github.com/Ch-LZ/tidyhebo. Data and code to reproduce our results can be obtained from https://github.com/Ch-LZ/tidyhebo_results. Data and code for the Olympus benchmark were obtained from https://github.com/aspuru-guzik-group/olympus. Data and code for evaluation on NiaH problems were obtained from https://github.com/PV-Lab/ZoMBI-Hop. Code for the Bayesmark benchmark was updated to match current library versions and is available at https://github.com/Ch-LZ/bayesmark.

# Funding

This work was supported by the Ministry of Economic Development of the Russian Federation in accordance with the subsidy agreement (agreement identifier 000000C313925P4H0002; grant No 139-15-2025-012).

# Conflict of interest

The authors declare that they have no known competing financial interests or personal relationships that could have appeared to influence the work reported in this paper.

# References

[1] Liang Q, Gongora AE, Ren Z, et al. Benchmarking the performance of Bayesian optimization across multiple experimental materials science domains. npj Comput Mater 2021;7:188. https://doi.org/10.1038/s41524-021-00656-9.

[2] Lei B, Kirk TQ, Bhattacharya A, et al. Bayesian optimization with adaptive surrogate models for automated experimental design. npj Comput Mater 2021;7:194. https://doi.org/10.1038/s41524-021-00662-x.

[3] Heifetz A, editor. High Performance Computing for Drug Discovery and Biomedicine. vol. 2716. New York, NY: Springer US; 2024. https://doi.org/10.1007/978-1-0716-3449-3.

[4] Shields BJ, Stevens J, Li J, et al. Bayesian reaction optimization as a tool for chemical synthesis. Nature 2021;590:89–96. https://doi.org/10.1038/s41586-021-03213-y.

[5] Wang K, Dowling AW. Bayesian optimization for chemical products and functional materials. Current Opinion in Chemical Engineering 2022;36:100728. https://doi.org/10.1016/j.coche.2021.100728.

[6] Roussel R, Edelen AL, Boltz T, et al. Bayesian optimization algorithms for accelerator physics. Phys Rev Accel Beams 2024;27:084801. https://doi.org/10.1103/PhysRevAccelBeams.27.084801.

[7] Wu Y, Walsh A, Ganose AM. Race to the bottom: Bayesian optimisation for chemical problems. Digital Discovery 2024;3:1086–100. https://doi.org/10.1039/D3DD00234A.

[8] Tellinghuisen J. Weighted least-squares in calibration: What difference does it make? Analyst 2007;132:536. https://doi.org/10.1039/b701696d.

[9] Jain RB. Comparison of three weighting schemes in weighted regression analysis for use in a chemistry laboratory. Clinica Chimica Acta 2010;411:270–9. https://doi.org/10.1016/j.cca.2009.11.021.

[10] Malz F, Jancke H. Validation of quantitative NMR. Journal of Pharmaceutical and Biomedical Analysis 2005;38:813–23. https://doi.org/10.1016/j.jpba.2005.01.043.

[11] Rasmussen CE, Williams CKI. Gaussian processes for machine learning. 3. print. Cambridge, MA: MIT Press; 2006. https://doi.org/10.7551/mitpress/3206.001.0001.

[12] Cowen-Rivers AI, Lyu W, Tutunov R, et al. HEBO: Pushing The Limits of Sample-Efficient Hyper-parameter Optimisation. Journal of Artificial Intelligence Research 2022;74:1269–349. https://doi.org/10.1613/jair.1.13643.

[13] Hebbal A, Brevault L, Balesdent M, et al. Bayesian Optimization using Deep Gaussian Processes 2019. https://doi.org/10.48550/arXiv.1905.03350.

[14] Marmin S, Ginsbourger D, Baccou J, et al. Warped Gaussian Processes and Derivative-Based Sequential Designs for Functions with Heterogeneous Variations. SIAM/ASA J Uncertainty Quantification 2018;6:991–1018. https://doi.org/10.1137/17M1129179.

[15] Cheng L, Ramchandran S, Vatanen T, et al. An additive Gaussian process regression model for interpretable non-parametric analysis of longitudinal data. Nat Commun 2019;10:1798. https://doi.org/10.1038/s41467-019-09785-8.

[16] Ta S, Liu T. Data-Driven Multi-Objective Optimization of an Industrial Hydrotreater Using a Multi-Output Gaussian Process Surrogate. Ind Eng Chem Res 2026;65:12743–60. https://doi.org/10.1021/acs.iecr.6c00111.

[17] Gramacy RB. Surrogates: Gaussian process modeling, design, and optimization for the applied sciences. Chapman and Hall/CRC; 2020.

[18] Khan A, Cowen-Rivers AI, Grosnit A, et al. Toward real-world automated antibody design with combinatorial Bayesian optimization. Cell Reports Methods 2023;3:100374. https://doi.org/10.1016/j.crmeth.2022.100374.
[19] Siemenn AE, Ren Z, Li Q, et al. Fast Bayesian optimization of Needle-in-a-Haystack problems using zooming memory-based initialization (ZoMBI). npj Comput Mater 2023;9:79. https://doi.org/10.1038/s41524-023-01048-x.
[20] Turner R, Eriksson D, McCourt M, et al. Bayesian Optimization is Superior to Random Search for Machine Learning Hyperparameter Tuning: Analysis of the Black-Box Optimization Challenge 2020. Proceedings of the NeurIPS 2020 Competition and Demonstration Track, vol. 133, PMLR; 2021, p. 3–26. https://doi.org/10.48550/arXiv.2104.10201.
[21] Balandat M, Karrer B, Jiang D, et al. BoTorch: A Framework for Efficient Monte-Carlo Bayesian Optimization. Advances in Neural Information Processing Systems, vol. 33, Curran Associates, Inc.; 2020, p. 21524–38. https://doi.org/10.48550/arXiv.1910.06403.
[22] Häse F, Aldeghi M, Hickman RJ, et al. Olympus: a benchmarking framework for noisy optimization and experiment planning. Mach Learn: Sci Technol 2021;2:035021. https://doi.org/10.1088/2632-2153/abedc8.
[23] Hickman R, Parakh P, Cheng A, et al. Olympus, enhanced: benchmarking mixed-parameter and multi-objective optimization in chemistry and materials science. ChemRxiv; 2023. https://doi.org/10.26434/chemrxiv-2023-74w8d.
[24] Ament S, Daulton S, Eriksson D, et al. Unexpected Improvements to Expected Improvement for Bayesian Optimization. Advances in Neural Information Processing Systems, vol. 36, Curran Associates, Inc.; 2023, p. 20577–612. https://doi.org/10.48550/arXiv.2310.20708.
[25] Jamil M, Yang X-S. A Literature Survey of Benchmark Functions For Global Optimization Problems. IJMMNO 2013;4:150. https://doi.org/10.1504/IJMMNO.2013.055204.
[26] Akiba T, Sano S, Yanase T, et al. Optuna: A Next-generation Hyperparameter Optimization Framework. Proceedings of the 25th ACM SIGKDD International Conference on Knowledge Discovery & Data Mining, New York, NY, USA: Association for Computing Machinery; 2019, p. 2623–31. https://doi.org/10.1145/3292500.3330701.
[27] Virtanen P, Gommers R, Oliphant TE, et al. SciPy 1.0: fundamental algorithms for scientific computing in Python. Nat Methods 2020;17:261–72. https://doi.org/10.1038/s41592-019-0686-2.
[28] Storn R, Price K. Differential Evolution – A Simple and Efficient Heuristic for global Optimization over Continuous Spaces. Journal of Global Optimization 1997;11:341–59. https://doi.org/10.1023/A:1008202821328.
[29] Le Riche R, Picheny V. Revisiting Bayesian Optimization in the light of the COCO benchmark. Struct Multidisc Optim 2021;64:3063–87. https://doi.org/10.1007/s00158-021-02977-1.
[30] Scyphers ME, Missik JEC, Kujawa H, et al. Bayesian Optimization for Anything (BOA): An open-source framework for accessible, user-friendly Bayesian optimization. Environmental Modelling & Software 2024;182:106191. https://doi.org/10.1016/j.envsoft.2024.106191.
[31] Snoek J, Larochelle H, Adams RP. Practical Bayesian Optimization of Machine Learning Algorithms. Advances in Neural Information Processing Systems, vol. 25, Curran Associates, Inc.; 2012. https://doi.org/10.48550/arXiv.1206.2944.
[32] Hvarfner C, Hellsten EO, Nardi L. Vanilla Bayesian Optimization Performs Great in High Dimensions. Proceedings of the 41st International Conference on Machine Learning, PMLR; 2024, p. 20793–817. https://doi.org/10.48550/arXiv.2402.02229.
[33] Grosnit A, Lyu W, Cowen-Rivers A, et al. huawei-noah/HEBO: v0.3.4 2022. https://doi.org/10.5281/zenodo.7344859.

[34] Box GEP, Cox DR. An Analysis of Transformations. Journal of the Royal Statistical Society Series B: Statistical Methodology 1964;26:211–43. https://doi.org/10.1111/j.2517-6161.1964.tb00553.x.

[35] Sakia RM. The Box-Cox Transformation Technique: A Review. The Statistician 1992;41:169. https://doi.org/10.2307/2348250.

[36] Yeo I-K, Johnson RA. A new family of power transformations to improve normality or symmetry. Biometrika 2000;87:954–9. https://doi.org/10.1093/biomet/87.4.954.

[37] Snelson E, Rasmussen CE, Ghahramani Z. Warped Gaussian Processes. Advances in Neural Information Processing Systems, vol. 16, MIT Press; 2003, p. 337–44.

[38] Astudillo R, Frazier PI. Bayesian Optimization of Composite Functions. Proceedings of the 36th International Conference on Machine Learning, PMLR; 2019, p. 354–63. https://doi.org/10.48550/arXiv.1906.01537.

[39] Blank J, Deb K. Pymoo: Multi-Objective Optimization in Python. IEEE Access 2020;8:89497–509. https://doi.org/10.1109/ACCESS.2020.2990567.

[40] Deb K, Jain H. An Evolutionary Many-Objective Optimization Algorithm Using Reference-Point-Based Nondominated Sorting Approach, Part I: Solving Problems With Box Constraints. IEEE Trans Evol Computat 2014;18:577–601. https://doi.org/10.1109/TEVC.2013.2281535.

[41] Deb K, Pratap A, Agarwal S, et al. A fast and elitist multiobjective genetic algorithm: NSGA-II. IEEE Trans Evol Computat 2002;6:182–97. https://doi.org/10.1109/4235.996017.

# Supplementary materials

## tidyHEBO: Improving the Robustness of Heteroskedastic Evolutionary Bayesian Optimization

Zhukov L.A.[1,2], Shaburova E.V.[2], Antonets D.V.[1,2]
[1] *AI Center MSU, Lomonosov Moscow State University, Moscow, Russia*
[2] *MSU Institute for Artificial Intelligence, Lomonosov Moscow State University, Moscow, Russia*

## Table of contents

# Supplementary Note 1 | Joint training objective derivation

We estimate the Yeo–Johnson parameter λ jointly with the GP hyperparameters by maximum a posteriori (MAP) optimization of a marginal log-posterior objective.

Let y denote the vector of observed responses, f the latent GP values on the transformed response scale, θ the GP parameters, and λ the transformation parameter. With independent priors on θ and λ, the joint density is

$$p(y, f, \theta, \lambda | X) \; = \; p(y|f, X, \theta, \lambda) \cdot p(f|X, \theta) \cdot p(\theta) \cdot p(\lambda)$$

Define $z = \psi_\lambda(y)$, where $\psi_\lambda(y_i) = T_\lambda(y_i/s_y)$ and $s_y$ is held fixed during a given model fit. Expressing the observation density on the original response scale gives

$$p(y, \mathbf{f}, \theta, \lambda \mid X) = p(z \mid \mathbf{f}, X, \theta)\, p(\mathbf{f} \mid X, \theta)\, \left|\det J_\psi(y, \lambda)\right|\, p(\theta)\, p(\lambda).$$

The transformation acts componentwise on the training observations. The Jacobian is therefore diagonal, and its determinant is the product of the componentwise derivatives.

For n observations, the Jacobian matrix is

$$J_\psi(y, \lambda) = \mathrm{diag}\left(\frac{1}{s_y} T'_\lambda(y_1/s_y), \ldots, \frac{1}{s_y} T'_\lambda(y_n/s_y)\right).$$

After integrating out the latent GP values and taking logarithms, the objective to maximize is

$$\mathcal{L}(y, \theta, \lambda) = \log p_{\mathrm{GP}}(z \mid X, \theta) + \log\left|\det J_\psi(y, \lambda)\right| + \log p(\theta) + \log p(\lambda).$$

Here, the first term is the GP marginal log-likelihood on the transformed targets, excluding parameter priors; the prior terms are written explicitly to avoid double counting. Because s_y is fixed during each fit, the Jacobian contribution −n log(s_y) is constant with respect to θ and λ.

$$(\hat{\theta}, \hat{\lambda}) = \arg\max_{\theta, \lambda} \mathcal{L}(y, \theta, \lambda).$$

This is a joint MAP estimate of λ and θ. The density identity assumes a valid invertible transformation on the support of the latent model; for λ outside [0, 2], the treatment of Gaussian mass outside the image of standard Yeo–Johnson must also be specified (see Supplementary Note 4).

## Supplementary Note 2 | Illustrative heteroskedastic example

Consider the following observation model:

$y(x) = f(x)\exp(\sigma\varepsilon - \sigma^2/2), \quad \varepsilon \sim \mathcal{N}(0,1).$

$\mathbb{E}[y(x) \mid x] = f(x).$

$\mathrm{Var}[y(x) \mid x] = f(x)^2(\exp(\sigma^2) - 1).$

The conditional variance is proportional to the squared mean.

When f(x) is strictly positive, a logarithmic transformation converts this construction to additive homoskedastic Gaussian noise:

$\log y(x) = \log f(x) + \sigma\varepsilon - \sigma^2/2.$

In this example, f(x) changes sign, so log(y) is not defined over the full domain. Yeo–Johnson accommodates signed responses, with λ estimated jointly with the GP. It need not exactly Gaussianize this signed multiplicative-noise model or make its transformed noise homoskedastic.

$f(x) = -0.5 + \sin(2\pi x).$

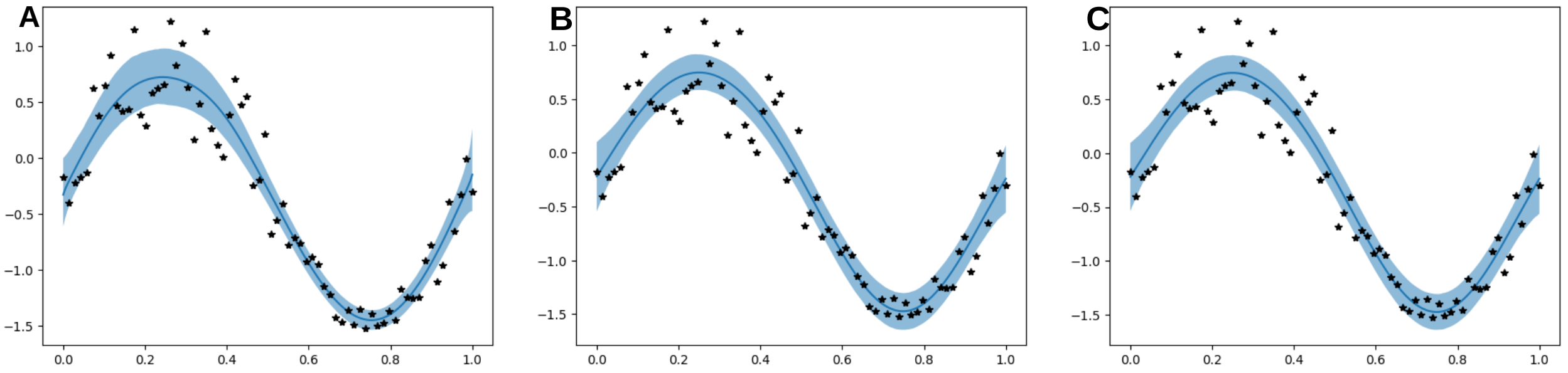

**Supplementary Figure 2.1 | Gaussian-process models fitted to the heteroskedastic example.** (A) The Yeo–Johnson transformation and GP are fitted jointly using one training objective. (B) The Yeo–Johnson transformation is fitted first, followed by the GP with the transformation fixed. (C) The GP is fitted without Yeo–Johnson output warping.

Supplementary Figure 1 illustrates how joint Yeo-Johnson-GP training changes predictive uncertainty relative to sequential transformation fitting and no output warping. In this synthetic example, joint training visually better follows the response-dependent dispersion induced by the data-generating process. The unwarped GP substantially misallocates predictive uncertainty between high- and low-response regions, whereas sequential Yeo-Johnson fitting gives an intermediate result. This illustration is not a quantitative calibration test.

A related illustration of GP behavior on a non-stationary objective is provided by the modified Xiong function in [1].

## Supplementary Table 1 | Performance metrics on the Ackley function

Lower values are better; bold and underlined values indicate the best and second-best results, respectively.

| Method | Median nAUC ↓ | CVaR_nAUC ↓ | IQR_nAUC ↓ |
|---|---|---|---|
| Optuna | 75.3 | <u>85.6</u> | <u>6.5</u> |
| Random search | 85.2 | 91.1 | **5.2** |
| HEBO | 78.7 | 90.6 | 38.4 |
| tidyHEBO | **25.5** | **74.0** | 17.9 |
| LogEI | <u>57.5</u> | 88.1 | 29.1 |

## Supplementary Table 2 | Performance metrics on the Hartmann function

Lower values are better; bold and underlined values indicate the best and second-best results, respectively.

| Method | Median nAUC ↓ | CVaR_nAUC ↓ | IQR_nAUC ↓ |
|---|---|---|---|
| Optuna | 34.4 | 55.3 | 10.0 |
| Random search | 51.3 | 62.1 | 13.3 |
| HEBO | **14.2** | **28.5** | <u>7.0</u> |
| tidyHEBO | <u>18.0</u> | <u>33.8</u> | **5.7** |
| LogEI | 20.4 | 56.7 | 15.0 |

## Supplementary Figure 2 | Task-level optimization performance and robustness across Olympus tasks

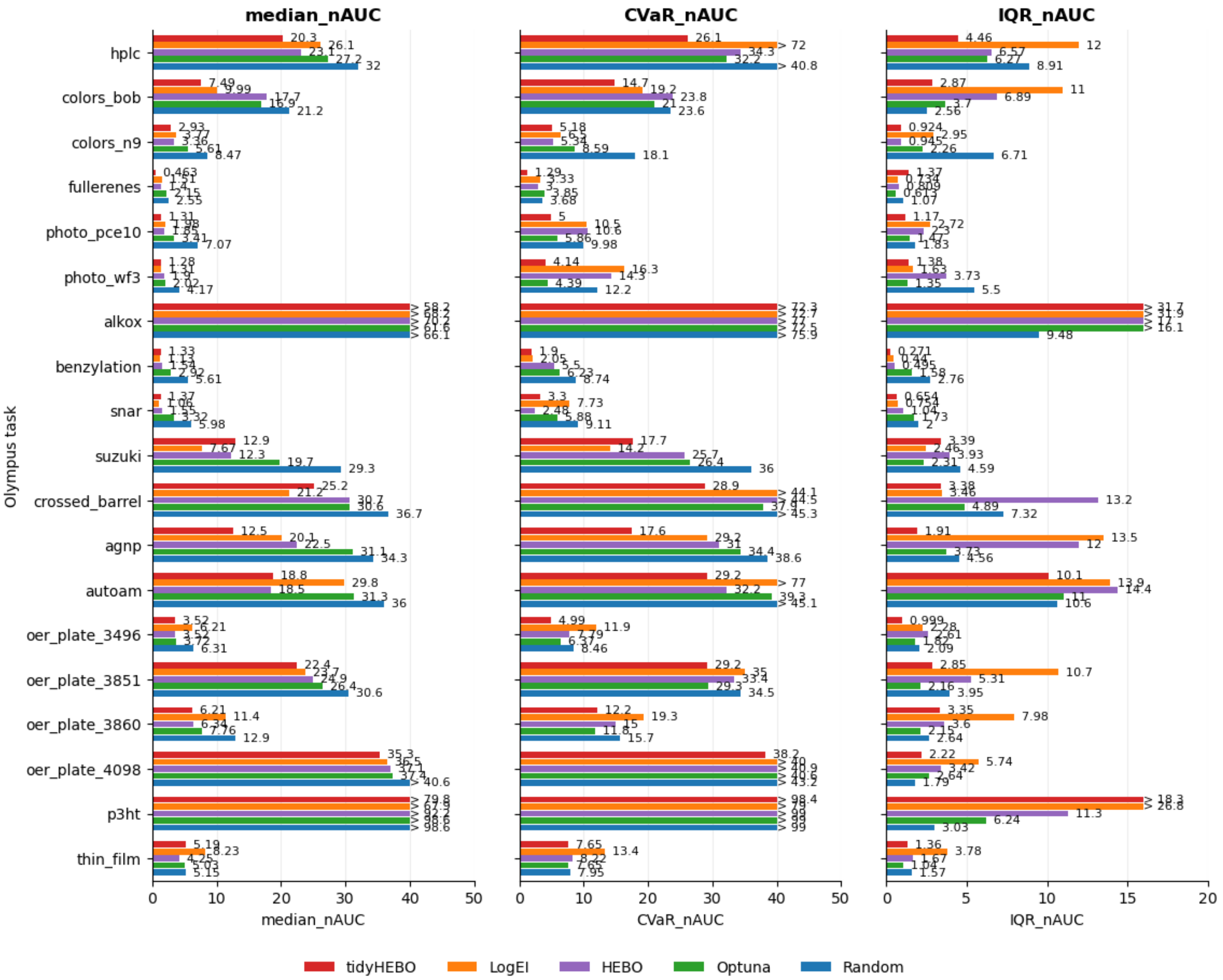

**Supplementary Figure 2 | Task-level optimization performance and robustness across Olympus tasks.** Median nAUC, CVaR_nAUC, and IQR_nAUC are shown for tidyHEBO, LogEI, HEBO, Optuna, and random search on each of the 19 single-objective Olympus tasks. Lower values indicate better performance for all metrics. Although objective values were mapped to a common min-max utility scale before nAUC calculation, absolute metric values still differ substantially across tasks, reflecting differences in optimization difficulty and response-surface structure.

## Supplementary Note 3 | Parametrization of simplex-constrained Olympus problems

Some Olympus benchmark problems contain k compositional variables constrained to the interior of the simplex $\delta^{k-1} = \{x \in \mathbb{R}^k: x_i > 0, \sum_{i=1}^{k} x_i = 1\}$. These variables are represented during optimization by k-1 auxiliary coordinates z using the additive log-ratio parametrization

$z_i = log\,(x_i/x_k), i = 1, \dots, k-1.$

The inverse map is

$$x_i = exp\,(z_i)\,/\,[1 + \sum_{j=1}^{k-1} exp\,(z_j)],$$
$$x_k = 1\,/\,[1 + \sum_{j=1}^{k-1} exp\,(z_j)],$$

Equivalently, $x = \mathrm{softmax}(z_1, \dots, z_{k-1}, 0)$. The unrestricted map is smooth and bijective between $\mathbb{R}^{k-1}$ and the interior of the k-component simplex. In the experiments, auxiliary coordinates were restricted to $[-2,2]^{k-1}$, so optimization covered only a subset of that interior. In particular, $e^{-2} \leq x_i/x_k \leq e^2$ for each i < k; boundary compositions with zero components were excluded.

## Supplementary Note 4 | Numerical evaluation of acquisitions after output warping

Acquisition functions in tidyHEBO are evaluated in the original objective space after inverse output transformation. This note describes the numerical integration used to compute these quantities efficiently and the relation between Gaussian quadrature and quasi-Monte Carlo estimation.

When output warping is active, the GP posterior is Gaussian on the transformed scale, $Z(x) \sim \mathcal{N}(\mu_z(x), \sigma_z^2(x))$, and the original-scale predictive variable is $Y(x) = s_y T_\lambda^{-1}[Z(x)]$, subject to the inverse-domain qualification below. EI averages $\max(Y(x) - y_{\mathrm{best}}, 0)$; UCB uses the original-scale mean and standard deviation, $\mathbb{E}[Y(x)] + \sqrt{\beta}\sqrt{\mathrm{Var}[Y(x)]}$. These quantities can be approximated by sampling Z, inverse-transforming the samples, and evaluating the relevant moments or improvement.

For sequential tidyHEBO optimization, the same one-dimensional integrals can instead be evaluated using Gaussian quadrature. Rather than drawing many samples, quadrature evaluates the integrand at a fixed set of points $z_j$ and combines the resulting values using predefined weights $w_j$: $\int f(z)p(z)dz \approx \sum_j w_j f(z_j)$.

Thus, quadrature can be viewed simply as another numerical method for evaluating the integral underlying an acquisition function: the function is evaluated at selected points, and the integral is approximated by their weighted sum. Unlike Monte Carlo integration, the quadrature calculation itself is deterministic.

For a strictly increasing transformation and a consistently transformed threshold, PI follows directly from the Gaussian latent posterior. In the sequential quadrature path, EI uses Gauss–Legendre quadrature and the moments required for UCB use Gauss–Hermite quadrature, with 512 nodes by default. The main text specifies this path for λ in [0, 2].

For EI, Gauss–Legendre quadrature integrates over the probability variable $p \in [\Phi(a), 1]$, where $a = \left[T_\lambda\left(y_{\text{best}}/s_y\right) - \mu_z\right]/\sigma_z$ for $\sigma_z > 0$. Here, Φ is the standard normal cumulative distribution function. Standard nodes on [−1, 1] and their weights are mapped to this interval. Each node is then mapped to $z = \mu_z + \sigma_z \Phi^{-1}(p)$ before inverse transformation and evaluation of the improvement. For zero posterior variance, EI reduces to the positive part of deterministic improvement.

When the one-dimensional quadrature path is not applicable, the described implementation uses Sobol quasi-Monte Carlo estimation with 4096 samples. Both procedures approximate the same acquisition criterion when applied to the same well-defined predictive distribution; changing the integration method alone does not resolve inverse-domain restrictions.

For standard Yeo–Johnson, the forward map has image $\mathbb{R}$ only for λ in [0, 2]. For λ < 0 its image is (−∞, −1/λ), and for λ > 2 it is (1/(2 − λ), ∞). Thus, for λ outside [0, 2], inverse transformation of an unrestricted Gaussian posterior is not defined everywhere. A complete specification must state how predictions outside this image are handled. If the latent distribution is truncated, its normalization and the corresponding changes to PI, EI, UCB, and the training likelihood must be included.

For a Gaussian predictive distribution on the objective scale, PI and EI have standard closed forms and UCB is computed directly from its mean and standard deviation. Nonlinear inverse warping generally makes the original-scale distribution non-Gaussian, so the standard Gaussian EI formula and latent Gaussian moments cannot be used directly for original-scale EI and UCB. PI remains analytic under a strictly increasing transformation with a consistently transformed threshold, provided the stated predictive distribution and its support are valid.

## Supplementary Note 5 | Details on benchmarking procedures

Unless stated otherwise, each optimizer was evaluated for 100 sequential function evaluations and each task was repeated 30 times from randomized initial conditions. Except for the Needle-in-a-Haystack (NiaH) results, mean trajectories were used for the primary optimization curves, with 95% bootstrap confidence intervals where shown. Reference optimizers were kept close to their default configurations: Optuna used the default TPE sampler, and LogEI used qLogExpectedImprovement with 4096 quasi-Monte Carlo samples. For sequential tidyHEBO, acquisition quantities after output warping were evaluated using the deterministic quadrature described in Supplementary Note 4 (512 nodes by default); Sobol quasi-Monte Carlo estimation with 4096 samples was used as a fallback when the one-dimensional quadrature expressions were not applicable.

**Evaluation on synthetic functions.** We used the Ackley and Hartmann implementations provided by BoTorch [2].

**Evaluation on Olympus.** Olympus [3,4] provides neural-network (NN) and Bayesian neural-network (BNN) emulators. We used the BNN variant for the 19 single-objective tasks evaluated in this study.

**Evaluation on NiaH problems.** Results for the ZoMBI methods were taken from the original study [5], and the random-forest emulators released with that work were used for the remaining evaluations. Each optimizer evaluated here was run 30 times on each task, with 100 function evaluations per run. Because the run-level outcomes are strongly skewed for these NiaH tasks, results are summarized by the median rather than the mean; confidence intervals are omitted from the corresponding figures for clarity.

**Evaluation on Bayesmark.** Each optimizer was given 100 function evaluations on all 108 problems, with the first four evaluations sampled randomly. One full 108-problem evaluation produced a Bayesmark leaderboard score [6]; this procedure was repeated 20 times per optimizer and the resulting scores were averaged. To ensure comparable and reproducible results, we used a reference baseline from the BBO Challenge [6]. The 95% confidence intervals shown in Figure 7 (main text) were estimated by bootstrap resampling of the 20 scores using 1000 resamples.

To preserve the composition of the historical Bayesmark benchmark under the scikit-learn versions used in this study [7], the Boston housing dataset used by the original benchmark was restored after its loader was removed from scikit-learn. The dataset was used solely to reproduce the historical benchmark composition.

## Supplementary References

[1] Hebbal A, Brevault L, Balesdent M, et al. Bayesian Optimization using Deep Gaussian Processes 2019. https://doi.org/10.48550/arXiv.1905.03350.

[2] Balandat M, Karrer B, Jiang D, et al. BoTorch: A Framework for Efficient Monte-Carlo Bayesian Optimization. Advances in Neural Information Processing Systems, vol. 33, Curran Associates, Inc.; 2020, p. 21524–38. https://doi.org/10.48550/arXiv.1910.06403.

[3] Hickman R, Parakh P, Cheng A, et al. Olympus, enhanced: benchmarking mixed-parameter and multi-objective optimization in chemistry and materials science. ChemRxiv; 2023. https://doi.org/10.26434/chemrxiv-2023-74w8d.

[4] Häse F, Aldeghi M, Hickman RJ, et al. Olympus: a benchmarking framework for noisy optimization and experiment planning. Mach Learn: Sci Technol 2021;2:035021. https://doi.org/10.1088/2632-2153/abedc8.

[5] Siemenn AE, Ren Z, Li Q, et al. Fast Bayesian optimization of Needle-in-a-Haystack problems using zooming memory-based initialization (ZoMBI). npj Comput Mater 2023;9:79. https://doi.org/10.1038/s41524-023-01048-x.

[6] Turner R, Eriksson D, McCourt M, et al. Bayesian Optimization is Superior to Random Search for Machine Learning Hyperparameter Tuning: Analysis of the Black-Box Optimization Challenge 2020. Proceedings of the NeurIPS 2020 Competition and Demonstration Track, vol. 133, PMLR; 2021, p. 3–26. https://doi.org/10.48550/arXiv.2104.10201.

[7] Pedregosa F, Varoquaux G, Gramfort A, et al. Scikit-learn: Machine Learning in Python. Journal of Machine Learning Research 2011;12:2825–30. https://doi.org/10.48550/arXiv.1201.0490.